\documentclass[a4paper]{wccm2020}
\usepackage{parskip}
\usepackage{pslatex}
\usepackage{graphicx}
\usepackage{amsmath}
\usepackage{amsfonts}
\usepackage{amssymb}
\usepackage{graphicx}
\usepackage{amsmath}
\usepackage{amsfonts}
\usepackage{amssymb}

\newcommand{\elAiv}{\ensuremath{\mathsf{A}_\mathsf{4}}}
\newcommand{\elEii}{\ensuremath{\mathsf{E}_\mathsf{2}}}
\newcommand{\elIii}{\ensuremath{\mathsf{I}_\mathsf{2}}}
\newcommand{\elMiii}{\ensuremath{\mathsf{M}_\mathsf{3}}}
\newcommand{\elNiii}{\ensuremath{\mathsf{N}_\mathsf{3}}}
\newcommand{\elPi}{\ensuremath{\mathsf{P}_\mathsf{1}}}
\newcommand{\elSi}{\ensuremath{\mathsf{S}_\mathsf{1}}}
\newcommand{\elSiv}{\ensuremath{\mathsf{S}_\mathsf{4}}}
\newcommand{\elUiv}{\ensuremath{\mathsf{U}_\mathsf{4}}}
\newcommand{\elsSIMU}{\elSi\elIii\elMiii\elUiv}
\newcommand{\formula}{\ensuremath{\mathnormal{\varphi}}}
\newcommand{\indvlx}{\ensuremath{\mathnormal{x}}}
\newcommand{\indvly}{\ensuremath{\mathnormal{y}}}

\newcommand{\object}{\ensuremath{\mathnormal{o}}}
\newcommand{\partof}{\ensuremath{\mathsf{P}}}
\newcommand{\pos}{\ensuremath{\mathnormal{\Diamond}}}
\newcommand{\represents}{\ensuremath{\mathsf{R}}}
\newcommand{\sign}{\ensuremath{\mathnormal{s}}}

\newcommand{\qq}[1]{``{#1}''}

\newcommand{\cf}{\textit{cf.}}
\newcommand{\eg}{\textit{e.g.}}
\newcommand{\etal}{\textit{et al.}}
\newcommand{\etc}{\textit{etc.}}
\newcommand{\ie}{\textit{i.e.}}

\title{SEMANTIC INTEROPERABILITY BASED ON THE EUROPEAN MATERIALS AND MODELLING ONTOLOGY AND ITS ONTOLOGICAL PARADIGM: MEREOSEMIOTICS}

\author{M.\ T.\ Horsch,$^{1}$ S.\ Chiacchiera,$^2$ B.\ Schembera,$^{1}$ M.\ A.\ Seaton,$^2$ and I.\ T.\ Todorov$^2$}

\heading{M.\ T.\ Horsch, S.\ Chiacchiera, B.\ Schembera, M.\ A.\ Seaton, I.\ T.\ Todorov}

\address{$^{1}$ High Performance Computing Center Stuttgart (HLRS)\\
Nobelstr.\ 19, 70569 Stuttgart, Germany\\
\{martin.horsch, bjoern.schembera\}@hlrs.de
\and
$^{2}$ UK Research and Innovation, STFC Daresbury Laboratory\\
Keckwick Ln, Daresbury, Cheshire WA4 4AD, United Kingdom\\
\{silvia.chiacchiera, michael.seaton, ilian.todorov\}@stfc.ac.uk}

\keywords{Research data infrastructure, semantic interoperability, top-level onto\-logies}

\abstract{The European Materials and Modelling Ontology (EMMO) has recently been
advanced in the computational molecular engineering and multiscale modelling communities as a top-level ontology,
aiming to support semantic interoperability and data integration solutions, \eg,
for research data infra\-structures.
The present work explores how top-level ontologies that are based on the same
paradigm -- the same set of fundamental postulates -- as the EMMO can be
applied to models of physical systems
and their use in computational engineering practice.
This paradigm, which combines mereology (in its extension as mereotopology)
and semiotics (following Peirce's approach), is here referred to as mereo\-semiotics.
Multiple conceivable ways of implementing mereosemiotics are compared, and
the design space consisting of the possible types of top-level ontologies
following this paradigm is characterized.}

\begin{document}

\section{INTRODUCTION} 

Semantic interoperability is the ability of multiple parties to exchange
information with a well-defined, mutually agreed meaning. This may be
codes that are coupled or linked to each other in a workflow or, \eg,
human or virtual agents that ingest, search, or extract
information as users of a research data infra\-structure.
Interoperability is called semantic whenever it is not simply based on a
common file format -- which would be syntactic interoperability -- but
on a more abstract formalization of the meaning of
data and metadata items, \ie, on the semantics of a common language; on this
basis, any serialization of the communicated content can be viably used as
long as its semantics has been defined clearly in terms of a mapping
from the employed syntax to
the representation of meaning in terms of that common language.
No FAIR data management, and hence no research data infrastructure that
adheres to the FAIR principles (findability, accessibility, interoperability,
and reusability), \cf~Bicarregui~\cite{Bicarregui16}, can be implemented without
such a common language~\cite{CGHHJLMPW18, Guizzardi20}. More generally,
any effort in simulation and data technology
that addresses communities as a whole, including many teams of developers,
cannot succeed without facilitating interoperability between the contributions
of these teams, be it in terms of software, data, or anything else that needs to
be integrated into a joint framework. For applications where it is
inevitable to connect a variety of components to each other
reliably, a single file format often emerges as a de facto standard, delivering
syntactic interoperability; in all other cases, even when there are only two
formats between which a conversion becomes necessary, it is unavoidable to
map them to each other in a way that is either implicitly or explictly
grounded in shared semantics. Making this
explicit requires more work, but delivers solutions that are more robust and
easy to extend to a greater number of parties. Whatever provides a definition of such an
underlying mutually agreed semantic space is called a semantic asset or also,
equivalently, a semantic artefact. The two most common approaches to
realizing this are hierarchical metadata standards, most typically based on XML/XSD
technology, where one element
acts as a container for others, and semantic-web technology
where resources (\ie, data and metadata items) are connected to each other
in a non-hierarchical way, yielding a knowledge graph; in the latter case,
an RDF schema or an OWL onto\-logy based on
description logic (DL) can be employed as a semantic asset,
while RDF triples are used to assert propositions concerning the data~\cite{AH11, BHLS17}.
  
Semantic assets are further distinguished by the level of abstraction at which
they operate: At one end, they can be very generic and remote from any concrete
application, as it is the case for top-level onto\-logies (also known as
foundational or upper onto\-logies), while other standards are domain-specific.
An ontology is a top-level ontology if it claims to formalize all that exists
and all ways in which objects can be related to each other. Both technically and
by the logic of the approach itself, it is by no means necessary to use a
top-level onto\-logy in order to realize semantic interoperability.
Instead, top-level onto\-logies play a role that is analogous to that of widespread
libraries and APIs in software engineering; they can make coding efforts less
redundant by providing a reliable generic basis for dedicated developments that are more
specific, and wherever they are employed by multiple components, it becomes easier
to integrate them into a single coherent solution. However,
the use cases that drive metadata standardization and hence the development
of semantic assets almost always reside within a comparably narrow domain of
knowledge; accordingly, the semantic space to be characterized, \ie, the expressive capacity
of the agreed language in terms of what sort of propositions can be formulated,
pertains to a specific field that constitutes only a very small fraction of all
that which can be known or asserted in general. 

The present line of work, in
particular, is concerned with use cases from molecular and multiscale modelling
and simulation as well as research data management concerning
thermodynamic, mechanical, and kinetic properties that are of interest to
chemical and process engineering.
Gygli and Pleiss~\cite{GP20} argue that despite significant efforts at
making molecular simulation data and metadata FAIR, it is still mainly due
to the lack of a recognized \qq{community wide standard} for semantic
interoperability, or of an \qq{MDML} (molecular dynamics markup language), that
it is hard to communicate molecular models and simulation
workflows \qq{from one molecular dynamics package to another.}
According to Edwards~\etal~\cite{EMBBB11}, metadata can serve as a key contributor to
diminish \qq{science friction} and enable interoperability. Semantic heterogeneity,
however, can contribute to \qq{metadata friction} which Edwards~\etal~\cite{EMBBB11} define
as the effort required to use metadata as a product (\eg, the effort to
tag data with metadata). In our sense, metadata friction would also occur when a
metadata model or semantic asset lacks recognition as a community wide
standard, and it would include the effort required to achieve a broad community agreement. 
Since the emergence of semantic technology,
there have been many endeavours to develop metadata standards for
modelling and simulation~\cite{TPT10}. Incidentally, a molecular
simulation markup language (MSML) had been in
existence since 2014, \cf~Grunzke~\etal~\cite{GBGHKBDPS5MJNAK14}, and a markup-language
based standard for engineering metadata (EngMeta) had been released by the
University of Stuttgart~\cite{SI19, SI20}
-- but as it is clear from the assessment by Gygli and Pleiss~\cite{GP20}, it
is broad community uptake that counts,
not the development of a semantic asset as such.
Similarly, for the development of a molecular model database~\cite{SHVH19a},
substantial work had to be done to clarify the employed
nomenclature; not because there was no clear understanding of the terminology,
but precisely due to the absence of a broad community agreement.
In this and many similar cases, including work done
by a series of projects funded from the Horizon 2020 research
and innovation programme such as VIMMP~\cite{HNBC3ELNSSTVC20, HCSTSLABMGKSFBSC20, HCSTSKK20, HCCS21} and
EMMC-CSA~\cite{EMMC17, Cenelec18, HAKBGHS19},
the \textit{Review of Materials Modelling} by de
Baas~\cite{DeBaas17} has been playing the role of an underlying foundational element.

Ontologies targeted at
simulation-based engineering include OntoCAPE~\cite{MWM08, MMWY10},
PhysSys \cite{BAT97} and, more recently,
the Physics-based Simulation Ontology (PSO)~\cite{CB19},
the Ontology for Simulation, Modelling, and
Optimization (OSMO)~\cite{HNBC3ELNSSTVC20, HCSTSLABMGKSFBSC20, HCSTSKK20, HCCS21},
and the European Materials and Modelling Onto\-logy~(EMMO),
a top-level ontology developed by Ghedini~\etal~\cite{url-emmo},
\cf~Goldbeck~\etal~\cite{GGHSF19} and Francisco Morgado~\etal~\cite{FGGHSFD20},
motivating the present discussion.
Among the metadata standards mentioned above, the EMMO is unique in that
it is a top-level ontology: It operates at the highest
level of abstraction and hence constitutes a contribution to the philosophy of
modelling and simulation in its own right. Moreover, the EMMO is also
particularly innovative as a top-level ontology by following a novel approach that raises
non-trivial issues and challenges when users attempt to apply it to concrete scenarios.
Some of these issues will be examined here with respect
to \textit{mereosemiotics, the paradigm of the EMMO}, understood to consist in a series of postulates
that underly not merely the EMMO itself, but its philosophical approach as such, of which
there can be a variety of implementations that would all face similar challenges.
As Francisco Morgado~\etal~\cite{FGGHSFD20} summarize, the EMMO is \qq{the top
and middle level onto\-logy developed by the EMMC} and, at the same time,
pursues domain-specific targets by aiming to become \qq{the only materials
science onto\-logy able to explicitly capture all granularity levels of
description of processes and materials.} Accordingly, in addition to its top-level
part, the EMMO also contains modules that have
the status of domain ontologies. These modules, however, are not the focus of
the present discussion which concerns \qq{metaontology} in the sense
suggested by Berto and Plebani~\cite{BP15}, \ie, it deals with design choices that need
to be made when a top-level ontology is developed. For a discussion of the
related, but distinct problem of making appropriate use of the existing
domain-specific metadata standards in materials modelling, the reader is referred to
an article dedicated to that topic~\cite{HFGIKS21}.

The present work therefore deals with the ontological task of characterizing
what models and simulations \textit{are} and how to describe this formally
in a coherent way at an abstract level~\cite{TPT10}. It is structured as follows:
In Section~\ref{sec:mereosemiotics}, those aspects from the approach of the EMMO
are isolated that are most fundamental and can thus be seen as constituting
a philosophical or ontological paradigm, here called mereo\-semiotics.
Section~\ref{sec:practical} introduces three challenges that any ontology
implementing this paradigm needs to address when it is applied to scenarios
from engineering data technology, and Section~\ref{sec:variations} discusses four
design choices to be made top-level ontology developers.
Section~\ref{sec:conclusion} concludes this article
by characterizing the top-level ontology design space that is accessible
to ontologies that implement mereosemiotics.

\section{TOP-LEVEL ONTOLOGIES AND MEREOSEMIOTICS}
\label{sec:mereosemiotics}

Formal onto\-logy (particularly, metaonto\-logy~\cite{BP15}) may appear rather
philosophical and abstract, and rightly so,
but it can nonetheless deliver actual technical support to
successful FAIR data management based on onto\-logy-driven semantic technology.
As Guizzardi states, \qq{the opposite of onto\-logy is not non-onto\-logy,
but just bad onto\-logy}~\cite{Guizzardi20}; a well-designed top-level onto\-logy
ensures that concepts and relations are integrated into a sound and coherent
formalization. Unsurprisingly, a broad spectrum of solutions for this problem are
available. The reader is referred to a recent
UKCIH survey~\cite{PMCLSW20} where the existing top-level ontologies
are discussed in detail. Many of them precede the EMMO by a long
time; \eg, in 2003, Masolo~\etal~\cite{MBGGO03}, conducted a systematic
comparison of the BFO, DOLCE, and OCHRE, and at a 2006 \qq{summit,}
a \textit{Joint Communiqu\'e}~\cite{OCRSSWY06} was issued wherein a group of developers
representing the BFO, DOLCE, GUM, ISO 15926, ISO 18629, OpenCyc, and SUMO
agreed to henceforth act as \qq{upper ontology custodians.} A comparison of how
top-level ontologies deal with temporal phenomena, focusing on the BFO~\cite{ASS15},
DOLCE~\cite{BM10}, and the GFO~\cite{Herre10}, was published by Galton~\cite{Galton18}.
With the EMMO, another stone has been added to this mosaic.
Most of the pre-existing top-level ontologies distinguish
between continuants and occurrents, following Johnson~\cite{Johnson24};
the EMMO, however, is based on a novel approach that combines
4D mereotopology and Peircean semiotics: Mereosemiotics.

The term \textit{mereosemiotics} is employed here for
a paradigm that is constituted by the following core tenets:
\begin{itemize}
   \item{} Mereology, \ie, 
      one of the fundamental relations is stated to be
      (improper) spatiotemporal parthood $\partof$ such that $\indvlx~\partof~\indvly$ means
      \qq{$\indvlx$ is (an improper) part of $\indvly$.}
      Mereology can be extended to mereo\-topology
      by including an irreducible predicate for connectedness~\cite{AV95, Varzi96, Smith96, SV00}.
   \item{} Semiotics following Peirce, by which representamina (signs)
      engage in a dialectical relationship with represented objects through a
      process the ele\-mentary steps of which are conceptualized as triads~\cite{Peirce55, Peirce91, Zeman77, Short07};
      representation $\represents$, where $\sign~\represents~\object$ would
      mean \qq{the representamen $\sign$ represents the object $\object$, its referent}
      always generates a third ele\-ment when it occurs in practice.
      By semiosis, in particular, an interpretant $\sign'$
      is created as a new representamen for the same object.
   \item{} Nominalism, by which only concrete things exist; ontological commitment
      exclusively extends to individuals (\ie, in other words, objects, instances, or concreta).
      Quantification cannot be applied to concepts (\ie, in other words, universals, classes, or abstracta),
      only to individuals, and only individuals, not concepts, can instantiate a concept.
      This is in line with the OWL DL approach
      to formulating ontologies in a decidable fragment of first-order logic~\cite{BHLS17};
      however, it contrasts with Peirce's point of view, which is Platonist (or realist),
      \ie, non-nominalist.
   \item{} Physicalist materialism; \ie, only that exists which can be
      conceived of as being physically present either in actual reality or in a hypothetical
      scenario to which the same laws of physics apply.
   \item{} Spatiotemporal monism; all that exists are
      finite four-dimensional spacetime regions, \ie, objects with a
      non-zero extension in three spatial dimensions and the fourth dimension of time~\cite{Williams10}.
      Therein, monism means that all objects \textit{are} spacetime, as opposed to a dualist
      approach where objects would be \textit{in} spacetime so that there would
      actually be two kinds of entities: The container (spacetime) and the contained (object).
      Here, instead, an object is identified with its extension in spacetime.
      Similary, continuant-occurrent dualism is absent from this paradigm:
      Processes and the objects that participate in them all enjoy equal status as
      4D spatiotemporal regions.
   \item{} Semiotic monism; the postulates above already exclude mind-body
      dualism, since only physical objects exist and therefore mental processes
      are a kind of physical processes. Beyond this, the present approach also
      excludes sign-object dualism: There is no rigid distinction
      between things that can act as a representamen and things that can act as an
      referent. In Peirce's work there are many cases where real-world
      objects, for which there are signs, act as signs themselves, particularly as
      an \qq{index,} a kind of sign. Examples from Augustine,
      who also held this position, include the word \textit{sign}, which is itself
      a sign and can therefore act as its own referent~\cite{Augustinus98}, and a situation
      where \qq{smoke signifies fire} because we see
      smoke and know that there must be a fire~\cite{Augustinus02}.
      This contrasts with Saussurean semiotics where the roles of the signifier and
      the signified (and the referent, where applicable) are usually
      understood to be taken by entities from disjoint categories.
\end{itemize}
Mereosemiotics combines a materialist approach that
is well suitable for discussing materials and their properties
with semiotics grounded in a complex sign-object interaction. In this way, intricate scenarios
from modelling and simulation as well as their relation to experimental data can be captured.
Physica\-lism suggests a view of
operating with signs, and a view of thought in general,
as being constituted by processes that are simultaneously
logical, social, and material; much of Peirce's work can be read in this way.
To model something means to represent it. This point of view is, \eg, advanced
by Dur\'an~\cite{Duran18} who states that by analysing the way in which this
representation occurs, \qq{three kinds of models based on their representational capacity
emerge, namely, phenomenological models, models of data, and theore\-tical models.}
The Peircean approach can contribute such an analysis; in particular, it can
provide a formalization of representation relations and cognitive processes to be employed by
digital infra\-structures that deal with multiscale modelling and simulation of physical
systems, possibly in combination with experimental procedures and data. Thereby,
computer simulations are semioses, \ie, processes by which a representamen (namely, a model) is evaluated,
producing another representamen (a different model or a computed property, \ie, an interpretant),
and data management needs to take the material preconditions of this process
into account, since they constitute relevant metadata: Where was
the simulation done, how were the input and the output
stored and exchanged, \etc? Any research data infrastructure that aims at making
models and simulation results FAIR must account for such metadata~\cite{SD19}.
Similarly, metadata on the provenance of sensory data are
crucial for integrating \qq{field devices 4.0} into model-driven process control~\cite{Maiwald20}.
In view of this, such an approach seems to be well
suited for its purpose.

\section{PRACTICAL CHALLENGES}
\label{sec:practical}

Mereosemiotics permits
describing modelling and simulation as semiosis. In this way,
central concepts from simulation-based engineering data technology can be
given a fundamental function in terms of
entities defined in a top-level ontology;
further domain ontologies can then be aligned with the top-level ontology,
yielding a data infra\-structure where models, data, and services can be integrated in
accordance with the principles of FAIR data stewardship~\cite{Bicarregui16, CGHHJLMPW18, Guizzardi20, SD19}.
However, the core tenets of this ontological paradigm also raise issues that
are not necessarily trivial to address. Three examples are given below:

\subsection{Mereotopology and collectives}

Mereotopology permits the distinction between spatiotemporal objects
that are connected components (called \qq{items} by the EMMO) as
well as mereotopological collectives, called \qq{collections} by
the EMMO, that are composed of multiple connected components.
As discussed by Masolo~\etal~\cite{MVFBP20}, many kinds of
collective-like entities occur frequently in practice (and hence will
also appear on research data infrastructures) that do not straightforwardly
align with this categorization. This includes
\textit{all Nobel laureates}, which can act as the referent of a sign,
namely, the text \qq{all Nobel laureates;} but some Nobel laureates
met and shook hands during their lifetime, by which they are spatiotemporally
connected and belong to the same \qq{item,} whereas others did not. It
is hard to relate the entity \textit{all Nobel laureates}
to the individual people systematically by a single membership relation based on
mereo\-topology. On the other hand, the \qq{collection} concept is so generic that it also applies to
entities that do not really function as collectives at all, since their members
are very different from each other, such as \textit{the spatiotemporal
fusion (union) of Konrad Zuse and the coffee vending
machine on the first floor of the HLRS building}.

\subsection{Contingent and counterfactual phenomena}
\label{subsec:contingent}

It is inherent in the nature of modelling and simulation that it
does not limit itself to that which is actually present. Simulating something
that is already right in front of our eyes is the least interesting use of a
model; instead, simulations often concern scenarios that are \textit{contingent}
(\qq{should we run the process at 200 or 400 kPa?})
or \textit{counterfactual} (\qq{what if the weather after Chernobyl had been different?}).
Even practically \textit{impossible} scenarios are relevant, since simulations permit
considering idealized states and processes, inaccessible by experiment,
by which theories and higher-order models can be validated or parameterized.
A knowledge base supporting such use cases needs to be able to consider
a simulation of a system where the proposition $\formula$ holds
while simultaneously asserting that $\lnot \formula$ obtains in actual reality.
Similarly, it needs to be able to consider $\formula$ as one eligible option
during optimization and $\lnot \formula$ as another, without thereby
asserting the contradiction that both $\formula$
and $\lnot \formula$ simultaneously hold in the real world.

Moreover, as Kaminski~\cite{Kaminski15} points out, it is characteristic of
technological innovation to induce states of \textit{hypercontingency} where
it is contingent whether $\formula$ is contingent, or where it is possible that $\formula$
is possible. In model-driven engineering design this is not the exception, but
the rule: It may or may not be possible to actually build the nanosurface
tailored to a specific purpose that was obtained as a theoretical optimum from
a series of simulations. Obversely, if model checking for a discrete process
model does not succeed at proving conclusively that a given
requirement is satisfied under all circumstances, this does not necessarily mean that the
requirement \textit{can} be violated, it means that it \textit{could}.
Logically expressing this kind of hypercontingency needs to rely on modal pluralism~\cite{Fine02}, \eg, by
combining different types of possibility operators; in the present example,
if $\formula$ is a violation of the requirement, $\pos_1 \formula$ is the possibility
of a violation, and $\pos_0 \pos_1 \formula$ is the possibility of that
possibility, where $\pos_0$ and $\pos_1$ are not the same operator.

%
%

\subsection{Application to copying, input/output, and data transfer}
\label{subsec:semiotic}

%

Dealing with multiple copies of the same data or metadata item is one
of the most basic functions of data technology. It is a prerequisite for
any exploitation of semantic interoperability in practice.
But under what conditions are multiple signs (\eg, models or simulation results),
or multiple semioses (\eg, simulations) to be regarded
as the same, as equivalent, as similar enough, or as
manifestations of the same information content? How can
a unification or subsumption of multiple entities under a
shared identity be expressed when they occur in
different spacetime regions, \eg, on different computers?
Peirce defines two different
types of signs for this purpose;
``the word `the' will usually occur from fifteen to twenty-five times on a
page. It is in all these occurrences one and the same word, the same legisign.
Each single instance of it is a replica''~\cite{Peirce55}.
The \textit{legisign} is to be understood as a pattern or rule (or a law, hence the name)
that needs to be applied to assess whether something is a \textit{replica} of it.

\section{VARIATIONS OF THE PARADIGM}
\label{sec:variations}

On the basis of the core tenets of mereosemiotics,
it is possible to take a variety of perspectives with regard to
specific issues, some of which are explored below. In particular,
this concerns 1)~the relation between semiotics and
physicalism; 2)~the relation between triads and dyads in Peircean semiotics;
3)~the role of objects from fictional or counterfactual scenarios
and their participation in semiosis; 4)~the equivalence of multiple instances or copies
of the same symbol, data item, or simulation workflow.

\subsection{Relation between semiotics and physicalism}
\label{subsec:physicalism}

Two variants will be considered:
\begin{itemize}
   \item{} \textit{Semiotics is fundamental} (ele\-ment \elSi); \ie,
      it is not assumed that the
      relation of represented and representing entities during
      semiosis can be expressed straightforwardly in physical terms.
      This includes conceptualizations where representation/semiosis
      and spatio\-temporal parthood are both fundamental on an equal footing,
      or where symbolic reasoning is considered on the basis of
      formal logic only, without specifying any precise connection to
      the characterization of spacetime.


   \item{} \textit{Physical participation is fundamental} (ele\-ment \elPi);
      in this view (held by the EMMO), semiosis is a process, \ie, a spatio\-temporal
      region, in which the sign and the interpretant
      actually participate physically; arguably, this means that the representamen and the cognition
      in which it is involved are 4D regions that overlap spatiotemporally.
      In a \textit{perception}, \eg, a measurement or experiment, the represented object needs to be
      physically present as well. In the case of an
      \textit{interpretation}, the physical participation of
      the object in the process is not required; the representation is
      carried over from the previous step, ultimately pointing back to a
      perception where the object was present.
\end{itemize}
Peirce's semiotics contains aspects of both ele\-ments: It is fundamentally
non-nominalist insofar as it admits the existence of universals,
and universals cannot participate in a physical process in the same way as
individuals that are understood to be spatiotemporal regions. However,
Peirce requires a \qq{real causal connection} to obtain between
the sign and the object; for this purpose, he distinguishes between semioses where
the object is necessarily physically present and cases where the causal connection
between the sign and the object may be indirect. The latter include
references to hypothetical entities (permitted by
Peirce) which by their nature cannot overlap with (and thereby participate in)
a non-hypothetical process.

\subsection{Relation between triads and dyads}
\label{subsec:triad-dyad}

Two types of realizations of the paradigm can be distinguished
on the basis of treating the semiotic triad either as fundamentally irreducible
or as non-elementary and reducible to a dyadic
relation:
\begin{itemize}
   \item{} \textit{Irreducibility of the semiotic triad} (ele\-ment \elIii).
      Following Peirce, mere dyadic representation
      of the type $\sign~\represents~\object$ does not occur; representation
      can only be realized in combination with a third ele\-ment.
      Hence, the relations that are elementary to
      semiosis do not connect the sign directly to the object
      but, instead, the sign to the semiosis process, the object to
      the semiosis process, \etc


   \item{} \textit{Elementary dyadic representation} (ele\-ment \elEii).
      Obversely, the relation that is seen as elementary to semiosis is the dyadic
      one that directly connects the representamen and the referent,
      as it is the case with the relation \qq{hasSign} in the EMMO.
\end{itemize}

\subsection{Modal propositions}
\label{subsec:fact-fiction}

A rough classification might begin by differentiating ontologies that permit
making statements on that which is possible, impossible, contingent, \etc\ (\cf~Section~\ref{subsec:contingent}),
from ontologies that do not permit this:
\begin{itemize}
   \item{} \textit{Modal propositions can be expressed} (ele\-ment \elMiii).
      Top-level ontologies that account for modality can further be distinguished
      in a variety of ways; this includes whether they implement
      Kripke's notion of possible worlds~\cite{Kripke80} (or even \qq{impossible worlds}~\cite{BJ19})
      or reject that interpretation~\cite{Vetter11}, modal monism versus
      modal pluralism, and the distinction between
      contingentism, according to which the existence of an individual may be
      contingent, and necessitism, stating that
      everything (\ie, all that exists) necessarily exists and that
      modal operators can only be applied to relations~\cite{Williamson15}.


   \item{} \textit{No modal propositions can be expressed} (ele\-ment \elNiii).
      Mereosemiotics does not require the notions of possibility and necessity;
      a top-level ontology following this paradigm can restrict itself to describing
      how things are rather than how they could or must be. This is the case for the EMMO.
\end{itemize}
Peirce's semiotics explicitly permits referring to hypo\-thetical objects
through signs that are factually present; an ontology that aims at capturing
such scenarios therefore needs to be of the first type.

\subsection{Relation between multiple replicas}
\label{subsec:copies}

Possible approaches for implementing the replica-legisign formalism,
\cf~Section~\ref{subsec:semiotic}, may include:
\begin{itemize}
   \item{} \textit{Unification by universals} (ele\-ment \elUiv).
      Concreta can share a feature by partaking in the same
      abstractum. This is closest to Peirce's realism (\ie, Platonism). For an
      implementation that is compatible with nominalism, a variety
      of solutions can be conceived of; \eg, sets or collectives may be defined that
      contain individuals that are similar in a certain respect,
      or an equivalence relation may be employed to state that its subject and object
      are replicas of the same item.
      What these approaches have in common is that it is \textit{externally
      posited} that two items are similar: The strings ``the''
      and ``the'' are the same word because the knowledge base states it.
      It is not possible for one interpreter to recognize them as the
      same and for another to believe them to be different.


   \item{} \textit{Absence of unification} (ele\-ment \elAiv).
      If whatever exists is spacetime, and exists as spacetime,
      different regions of spacetime by definition cannot be the same;
      accordingly, ``the'' and ``the'' are just different physical
      objects, one of which is printed more to the left, while the
      other is printed more to the right. It is \textit{not posited} that they are similar.
      This corresponds to
      an ontology that strongly prioritizes mereology (or mereotopology) over semiotics,
      as it is the case for the EMMO.


   \item{} \textit{Unification by semiosis} (ele\-ment \elSiv).
      This solution proposes that assessing equivalence
      is a process of
      pattern matching, which is a semiosis. Thereby,
      the sign is a pattern, the object is the item that is
      matched against the pattern, and the interpretant is the
      outcome of the pattern matching process, which may be acceptance, rejection,
      or any other assessment of how the object matches the pattern.
      If two objects match the same pattern, they are equivalent in
      a certain respect; however, this is \textit{not externally posited}, but
      subject to evaluation, where multiple interpreters
      may disagree.
\end{itemize}

\section{CONCLUSION}
\label{sec:conclusion}

By combining ele\-ments from Sections~\ref{subsec:physicalism}
to~\ref{subsec:copies}, a variety of strategies can
be followed for the design of a top-level ontology and its application
to research data infrastructures in the engineering sciences.
The choices for each of the four points mentioned above can be made independently;
no combinations of ele\-ments appear to be absolutely
irreconcilable with each other. Therefore, the whole product space
\begin{equation*}
   \{\elPi,\, \elSi\} ~\times~ \{\elEii,\, \elIii\}
      ~\times~ \{\elMiii,\, \elNiii\} ~\times~ \{\elAiv,\, \elSiv,\, \elUiv\}
\end{equation*}
is accessible, providing a landscape of possible types of top-level ontologies
within the paradigm of mereosemiotics. Peirce
permits a reading that positions him comparably closely to the top-level ontology
type \elsSIMU, as far as such a claim may be upheld for any onto\-logy
that follows nominalism rather than Platonist realism. The EMMO opposes Peirce on
each of the four issues. However, the main purpose here
should not consist in an exegesis of Peirce's body of work,
since there may be very good reasons to deviate from his views; it remains to be explored
by future work to what extent the different types from the accessible
design space are suitable as a semantic foundation for
research data infrastructures.

\bigskip

\noindent
\textbf{Acknowledgment.} This article reports on a contribution to the Minisymposium on Ontology-Based Materials Modelling, Optimization, and Design Applied to Modelling Translation Services and Business Decision Support Systems (OBM-MODA-MTS-BDSS) at WCCM-ECCOMAS 2020, which was held in 2021. This work was supported by activities of the Innovation Centre for Process Data Technology (Inprodat e.V.), Kaisers\-lautern; the minisymposium was organized by N.\ A.\ Konchakova and P.\ Klein. The co-authors M.T.H.\ and B.S.\ acknowledge funding from the German Research Foundation (DFG) through the National Research Data Infrastructure for Catalysis-Related Sciences (NFDI4Cat) within the National Research Data Infrastructure (NFDI) programme of the Joint Science Conference (GWK).

\bibliography{emmo-wccm-eccomas}
\end{document}